\documentclass{article} 
\usepackage[dvipsnames]{xcolor}
\usepackage{iclr2021_conference,times}

\usepackage{pifont}

\usepackage{amsmath,amssymb,amsthm}
\usepackage{graphicx}
\usepackage{xspace}
\RequirePackage[dvips]{epsfig} 
\RequirePackage{float}
\RequirePackage{fancyhdr}
\RequirePackage{pstricks,pst-plot,psfrag}
\RequirePackage{multirow}
\RequirePackage{rotating}
\RequirePackage{units}
\usepackage{booktabs}
\usepackage{pgfplots}
\newlength\figureheight
\newlength\figurewidth
\pgfplotsset{compat=1.14}
\usepackage{caption}
\usepackage{array}
\usepackage{subcaption}
\usepackage{setspace}
\usepackage{enumitem}
\usepackage{caption}

\usepackage[titletoc,title]{appendix}

\usepackage{tabularx}
\newcolumntype{L}[1]{>{\raggedright\arraybackslash}p{#1}}
\newcolumntype{C}[1]{>{\centering\arraybackslash}p{#1}}
\newcolumntype{R}[1]{>{\raggedleft\arraybackslash}p{#1}}

\usepackage{etoolbox,siunitx}
\usepackage[super]{nth}
\usepackage{nicefrac}
\robustify\bfseries

\usepackage{amsmath,amsfonts,bm}

\def\eqref#1{equation~\ref{#1}}

\def\1{\bm{1}}

\DeclareMathAlphabet{\mathsfit}{\encodingdefault}{\sfdefault}{m}{sl}
\SetMathAlphabet{\mathsfit}{bold}{\encodingdefault}{\sfdefault}{bx}{n}

\newcommand{\E}{\mathbb{E}}

\usepackage{hyperref}
\usepackage{url}
\usepackage{blindtext}

\title{How to make Deep RL work in Practice}

\author{
Nirnai Rao$^{1*}$\qquad
Elie Aljalbout$^{1*}$\qquad
Axel Sauer$^{2,3*}$\qquad
Sami Haddadin$^{1}$\\
$^{1}$Technical University of Munich\\
$^{2}$Max Planck Institute for Intelligent Systems, T{\"u}bingen\\
$^{3}$University of T{\"u}bingen\\
\quad\quad$^{1}$\{\tt\small {firstname.lastname\}@tum.de} \quad
$^{2}$\{\tt\small {firstname.lastname\}@tue.mpg.de}
}

\begin{document}
\newcommand\blfootnote[1]{%
  \begingroup
  \renewcommand\thefootnote{}\footnote{#1}%
  \addtocounter{footnote}{-1}%
  \endgroup
}
\renewcommand*{\thefootnote}{\fnsymbol{footnote}}
\footnotetext[1]{Shared first authorship.}
\blfootnote{This work was done while Axel Sauer was with the Technical University of Munich.}
\renewcommand*{\thefootnote}{\arabic{footnote}}
\iclrfinalcopy
\maketitle

\begin{abstract}
In recent years, challenging control problems became solvable with deep reinforcement learning (RL). To use RL for large-scale real-world applications, a certain degree of reliability in their performance is necessary. Reported results of state-of-the-art algorithms are often difficult to reproduce. One reason for this is that certain implementation details influence the performance significantly. Commonly, these details are not highlighted as important techniques to achieve state-of-the-art performance. Additionally, techniques from supervised learning are often used by default but influence the algorithms in a reinforcement learning setting in different and not well-understood ways. In this paper, we investigate the influence of certain initialization, input normalization, and adaptive learning techniques on the performance of state-of-the-art RL algorithms. We suggest which of those techniques to use by default and highlight areas that could benefit from a solution tailored explicitly to RL. 
\end{abstract}

\section{Introduction}

Reinforcement learning (RL) is a computational approach to automate goal-directed learning of a policy by maximizing cumulative reward in an environment  \citep{sutton2018reinforcement}. In combination with Artificial Neural Networks as function approximators, RL can deal with high dimensional spaces, like images and depth data \citep{openaiSolvingRubikCube2019,kalashnikovQTOptScalableDeep2018}. In recent years, RL has shown impressive results in video games \citep{mnihPlayingAtariDeep} and control applications \citep{lillicrapContinuousControlDeep2015}. A lot of research effort went into improving the state-of-the-art methods, with the goal of large-scale real-world applications \citep{kalashnikovQTOptScalableDeep2018}. To achieve this goal, a certain degree of reliability is a requirement. 
Previous work has shown that reproducing results of state-of-the-art RL algorithms is very difficult, and the robustness of these algorithms can be brittle \citep{hendersonDeepReinforcementLearning2017}. Minor differences in implementation, can lead to a major change in performance \citep{hendersonDeepReinforcementLearning2017,pardoTimeLimitsReinforcement2017}. Furthermore, different codebases use different tricks to improve performance and to ensure stability. 

This paper aims to investigate implementation details and tricks, which have a significant impact on the performance of RL algorithms, validate their usefulness, and examine their effects. Our main contribution is to identify techniques that have the most impact on the overall performance. These techniques can be split into three categories: initialization schemes, input normalization, and adapting learning rates and gradients. For the evaluation, we chose commonly used algorithms on a set of control benchmark tasks.  Our experimental analysis aims to provide guidelines on which techniques to use and which to avoid. For the sake of reproducibility, we open-source the implementation of the algorithms and our evaluation procedures: \url{https://github.com/Nirnai/DeepRL}. 

\section{Related Work}
It has been established in recent years that RL research suffers regarding reproducibility and reusability  \citep{hendersonDeepReinforcementLearning2017,islamReproducibilityBenchmarkedDeep2017}. \cite{hendersonDeepReinforcementLearning2017} investigated how reproducibility is affected by different hyperparameters and the number of samples. They also examine the effect of environmental characteristics and different open-source codebases. These results can strongly vary based on different hyperparameters and codebases. The details in the codebases, which caused the divergent performances, were not further investigated. Our work analyses those details in-depth and explores their impact. 
Due to the algorithms' stochastic nature, a sufficient number of samples is necessary to get an insight into the performance of a population. 
Confidence intervals and significance tests are necessary to determine if the difference in populations is, in fact, significant.
  \citep{colasHitchhikerGuideStatistical2019,Khetarpal2018REEVALUATERI,colasHowManyRandom2018} propose new evaluation practices to improve upon some of those issues. \cite{colasHitchhikerGuideStatistical2019} investigated statistical significance tests based on the performance population of SAC and TD3 on the Half-Cheetah-v2 environment.   

Furthermore, the RL research community made steps towards providing benchmarks with a large variety of tasks  \citep{tassaDeepMindControlSuite2018,duanBenchmarkingDeepReinforcement,mnihPlayingAtariDeep}. Some of these benchmarks focus on continuous state and action spaces  \citep{tassaDeepMindControlSuite2018,duanBenchmarkingDeepReinforcement}, others focus on discrete state and action spaces  \citep{mnihPlayingAtariDeep}. For instance, the Deep Mind Control Suite introduces a unified reward structure, enabling robust performance measures across environments. 
Open source baseline implementations are publicly available\footnote[1]{\url{https://github.com/openai/baselines}}\footnote[2]{\url{https://github.com/openai/spinningup}}\footnote[3]{\url{https://github.com/rll/rllab}}. The details of those implementations vary strongly and different hand-engineered techniques are used across codebases. In this work, we analyze those details and their impact on learning behavior and final performance. We hope to provide insights into previously unquestioned defaults and build a better understanding of commonalities and disparities between different implementations.

\section{Common Implementation Details}
This section introduces the techniques that we found to impact learning behavior and performance the most. First, we discuss initialization methods for neural networks used in deep learning. Second, we introduce input normalization and its theoretical justification. Last, we investigate adaptive learning techniques, which either modify the learning rate or the gradients themselves.

\subsection{Initialization}
Initialization can determine whether an iterative algorithm, like RL, converges. If the algorithm converges, initialization could furthermore influence the learning speed and the quality of the solution. Designing principled initialization schemes for neural networks is not trivial because neural network optimization is still not fully understood. Thereby Deep RL also lacks such initialization schemes and, hence, commonly follows a heuristically motivated procedure. Common initialization schemes in supervised learning (SL) are Xavier-, Kaiming/He-, LeCun-, and Orthogonal initialization \citep{montavonNeuralNetworksTricks2012,glorotUnderstandingDifficultyTraining,heDelvingDeepRectifiers2015,saxeExactSolutionsNonlinear2013}. Xavier initialization is usually employed if activation functions of a neural network are symmetric (like sigmoid or tanh). 
It is common to use symmetric activation function for policy gradient methods like Trust Region Policy Optimization (TRPO) \citep{schulmanTrustRegionPolicy2015} and Proximal Policy Optimization (PPO) \citep{schulmanProximalPolicyOptimization2017}. If the activation function is non-symmetric, for example, ReLU and Kaiming initialization are preferred. This activation function is often used in Q-learning methods like Twin Delayed Deep Deterministic Policy Gradient (TD3) or Soft Actor-Critic (SAC). LeCun and Orthogonal initialization are applicable for any activation function.
These initialization methods are supposed to break the symmetry of a neural network and reduce the chance of exploding and vanishing gradients \citep{goodfellowDeepLearning2016,montavonNeuralNetworksTricks2012}.
In RL, initialization has an additional impact on exploration since it determines the initial distribution from which actions are sampled. This impacts policy gradient algorithms differently than Q-learning algorithms because they use differently parameterized policies. The initialization schemes used in different baselines are not consistent. The default in the OpenAI Baseline\footnotemark[1] implementation is the Orthogonal initialization, whereas OpenAI SpinningUp\footnotemark[2] uses Tensorflow's default, the Xavier initialization.    

\subsection{Input Normalization}
Deep Learning commonly uses input normalization to increase the learning speed \citep{montavonNeuralNetworksTricks2012}. Likewise, learning speed, and therefore sample efficiency, is a significant concern in RL. 
Input normalization ensures that the input data is distributed according to a standard normal distribution by transforming it with the following equation.
\begin{equation}
    \tilde{s} = \frac{s - \mu}{\sigma}
\end{equation} 
$s$ is the input state, $\tilde{s}$ is the normalized state, $\mu$ is the mean, and $\sigma$ the standard deviation. If the mean of the inputs is close to zero, it is easier and faster for weights to change their sign if necessary. Furthermore, rescaling the variance to one standard deviation balances out the rate at which the weights connected to the input nodes learn \citep{montavonNeuralNetworksTricks2012}. 
In SL, applying this transformation is straight forward since the entire dataset is available at the beginning. In RL, this is not the case. Furthermore, the distribution of the input variables is non-stationary. Running estimates of the distribution parameters are necessary to implement input normalization in an RL setting. Welford's algorithm \citep{welfordNoteMethodCalculating1962} is most commonly used to implement such a running estimation\footnotemark[1].
\begin{subequations}
    \begin{align}
        \mu_n &\leftarrow \mu_{n-1} + \frac{s_n-\mu_{n-1}}{n} \\
        \sigma^2_n &\leftarrow \sigma^2_{n-1} + \frac{ (s_n-\mu_{n-1})(s_n-\mu_n)}{n}
    \end{align}
    \label{eq:RunningMeanVar}%
\end{subequations} 
where $\mu_n$ and $\sigma_n$ are respectively the estimated mean and standard deviation after $n$ samples.
The parameter estimates are usually not very accurate at the beginning of the learning process. This is due to the limited number of samples available in the early stages of learning. Hence, this technique does not necessarily impact the performance in the same way as would be expected from SL. This technique is represented in some of the major codebases\footnotemark[1]\footnotemark[3] but is missing in some other\footnotemark[2]. 

\subsection{Adaptive Learning Techniques}
We refer to methods that modify the gradients or the learning rate during training by adaptive learning techniques. 
Adaptive learning is especially crucial for Policy Gradient methods since they suffer from very noisy gradients and, hence, can potentially take updates in directions that do not point towards the true gradient. PPO and TRPO reduce the impact of noisy gradients with trust regions. Since trust regions are not usually computed exactly, additional techniques are required to deal with approximation errors. The clipped objective in PPO, for example, follows the heuristic of keeping consecutive policies close to each other. However, it does not enforce a trust-region  \citep{ilyasAreDeepPolicy2018}. The desired property of small policy changes is often not fulfilled, making the vanilla implementation of PPO very unstable. 

This section introduces five techniques that are most commonly found in implementations of policy gradient algorithms\footnotemark[1]\footnotemark[2]\footnotemark[4].
These techniques are Learning Rate Schedules, Advantage Normalization, Gradient Clipping, KL-Stopping, and KL-Cutoff. They are often necessary to achieve state-of-the-art performance.

\textbf{Learning Rate Schedules\footnotemark[1] (LRS)} are used to decay the learning rate gradually throughout learning. This technique only appears in one baseline implementation\footnotemark[1], where it is used in combination with stochastic gradient descent to reduce noise. Especially in policy gradient methods, but not restricted to, it can help to manage noise. A popular choice for the schedule is a linear decay
\begin{equation}
    \alpha_t = \alpha_0 \cdot \left(1-\frac{t}{t_{\text{total}}}\right)
\end{equation}
where $\alpha_t$ is the learning rate at step $t$ and $t_{total}$ is the total number of steps.

\textbf{Advantage Normalization\footnote[1]{\url{https://github.com/openai/baselines}}\footnote[2]{\url{https://github.com/openai/spinningup}}\footnote[3]{\url{https://github.com/rll/rllab}}\footnote[4]{\url{https://github.com/joschu/modular_rl}} (AN)} transforms the advantages $\hat{A}_{\pi}$, used to compute the policy gradient, to have zero mean and unit variance. This acts as an adaptive learning rate heuristic and bounds the gradient variance \citep{tuckerMirageActionDependentBaselines2018}. It is one of the most commonly used techniques and is represented in all the baseline implementations we investigated\footnotemark[1]\footnotemark[2]\footnotemark[3]\footnotemark[4].
\begin{equation}
    \bar{A}_{\pi} = \frac{\hat{A}_{\pi} - \mu(\hat{A}_{\pi})}{\sigma(\hat{A}_{\pi})}
\end{equation}
\textbf{Gradient Clipping\footnotemark[1]} limits the maximum norm of the policy gradient. The gradient is modified according to the following update rule, until the threshold $\alpha$ is met.
\begin{equation}
\nabla_{\theta} J\left(\pi_{\theta}\right) \leftarrow \nabla_{\theta} J\left(\pi_{\theta}\right) \cdot \frac{\alpha}{\left\|\nabla_{\theta} J\left(\pi_{\theta}\right)\right\|^{2}}
\end{equation}
$\pi_{\theta}$ is the policy with parameters $\theta$, and $\nabla_{\theta} J\left(\pi_{\theta}\right)$ is the policy gradient. This technique limits the step size in parameter space, thereby avoiding very large gradient updates. It is commonly used to address the issue of exploding gradients and was only found in the OpenAI Baseline\footnotemark[1].

\textbf{KL-Stopping\footnotemark[2]} explicitly estimates the KL-Divergence after an update with the following equation
\begin{equation}
    \hat{D}_{\text{KL}} = \E\Big[ \log(\pi_{\theta_{\text{old}}}(a|s)) - \log(\pi_{\theta_{\text{new}}}(a|s)) \Big]
\end{equation}
If this value is larger than a threshold, optimization on the current data is stopped, and new data is collected. This technique directly addresses the fact that the clipped PPO objective does not guarantee small updates. The only baseline implementation using this technique calls it early stopping  \footnotemark[2].  In the following, we refer to this technique as KL stopping to avoid any confusion with early stopping from an SL context. 

\textbf{KL-Cutoff} also addresses the issues as mentioned earlier with the clipped objective. Instead of stopping optimization after violating the KL constraint, a correction term is added to the loss function of the next update, which is defined as follows.
\begin{equation}
    L = L^{\text{CLIP}} - \alpha \cdot (\hat{D}_{\text{KL}} > D_{\text{thr}}) \cdot (\hat{D}_{\text{KL}} - D_{\text{thr}})^2
\end{equation}
where $L^{\text{CLIP}}$ is the clipped objective and $D_{\text{thr}}$ is the KL constraint. After a violation of the constraint has occurred, this objective prioritizes the reduction of the KL-Divergence over the original objective. The resulting gradient step corrects the bad update and resumes optimization on the same data. This technique was used in the original implementation of PPO\footnotemark[3] but wasn't found in other baseline implementations\footnotemark[1]\footnotemark[2]\footnotemark[4].

\begin{figure}
    \centering
    \includegraphics[width=0.65\linewidth]{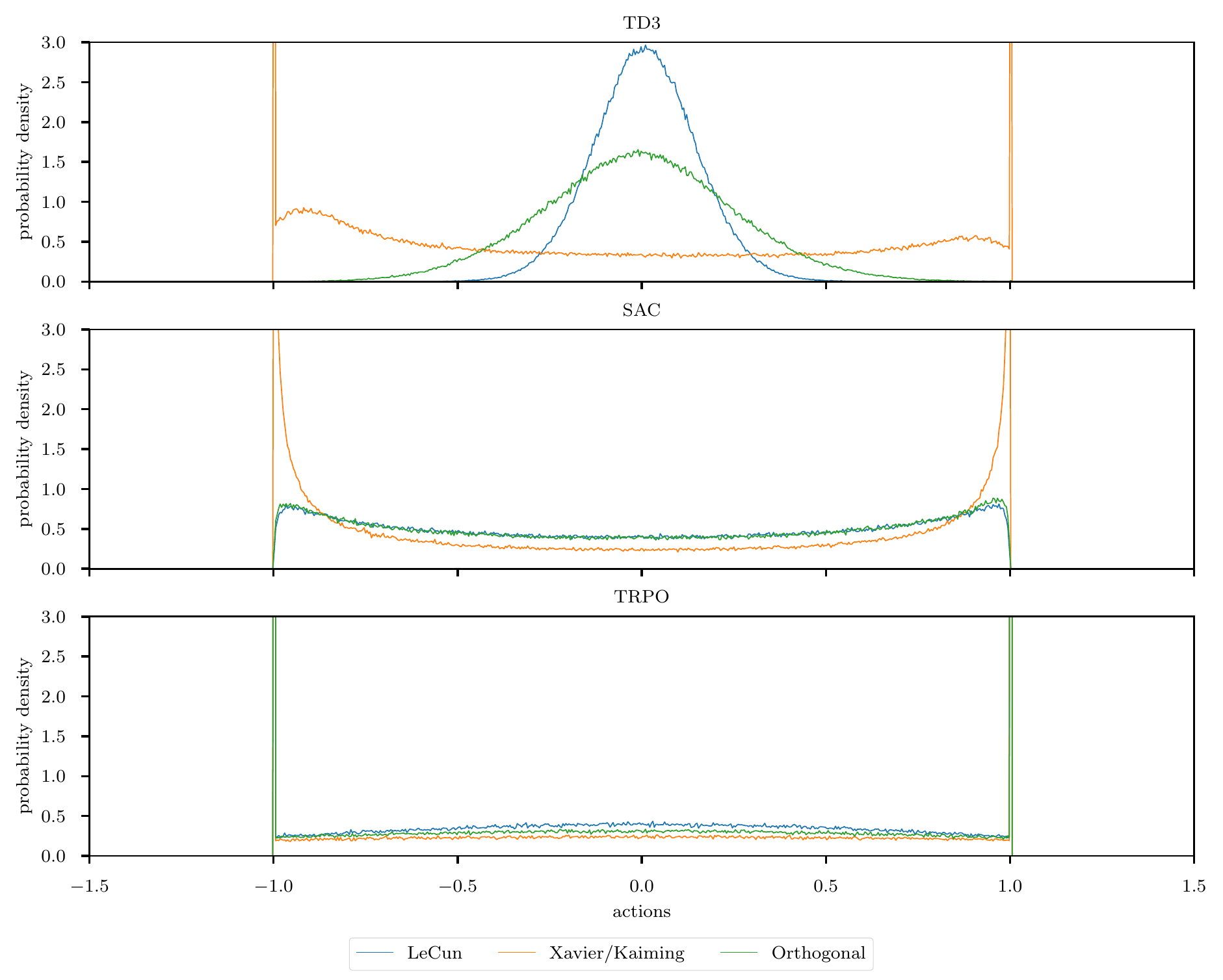}
    \caption{Probability density of actions selected under the initial policy for different
    initialization schemes in the cartpole environment for the swingup task.}
    \label{fig:histInit}
\end{figure}

\section{Experimental Analysis}
For the experimental analysis of initialization schemes and input normalization, we report results for TD3 and TRPO. These algorithms use fundamentally different parameterizations and represent the class of Q-learning methods and Policy Gradient methods, respectively. Additionally, we ran experiments on SAC and CGP to investigate Q-learning algorithms with differently parameterized policies. To facilitate the readability of the plots, we report additional results only if they provide additional insights. 
Adaptive learning techniques are evaluated on PPO since they are mainly used for policy gradient methods as they show particularly unstable learning behavior. All algorithms are implemented with the autograd library PyTorch \citep{pytorch}.
We run experiments on six continuous control tasks from the Deep Mind Control Suite, namely cartpole-balance, cartpole-swingup, acrobot-swingup, cheetah-run, hopper-hop, and walker-run. They cover linear and non-linear dynamics in low- and high-dimensional state and action spaces. Returns are collected in an offline manner and averaged over ten episodes. We run each algorithm for ten different seeds and report the mean performance.
This choice is common in recent publications \citep{fujimotoAddressingFunctionApproximation2018,simmons-edlerQLearningContinuousActions2019}. All observed effects have an effect size larger than two standard deviations, hence indicate strong evidence for these effects. 
Bootstrap confidence intervals are used to give a $95\%$ confidence level for the reported result.  

\subsection{Initialization}
Due to the underlying assumption that initialization influences differently parameterized policies in a different manner, we chose TD3, SAC, and TRPO.
TD3 uses a deterministic policy with additive noise. As for SAC, it uses a stochastic policy. Both policies bound their outputs by passing them through the $\tanh$ function. TRPO also uses a Gaussian policy, but without bounding the output. This results in a clipping of the actions by the environment. 
Note that Initialization only affects the mean value in the Gaussian policies. The standard deviations are initialized by the absolute value of the action limits. The standard deviation of the additive noise for the deterministic policy takes a value of $0.1$.

\begin{figure}
   \begin{minipage}[t]{0.48\linewidth}
     \centering
    \includegraphics[width=\columnwidth]{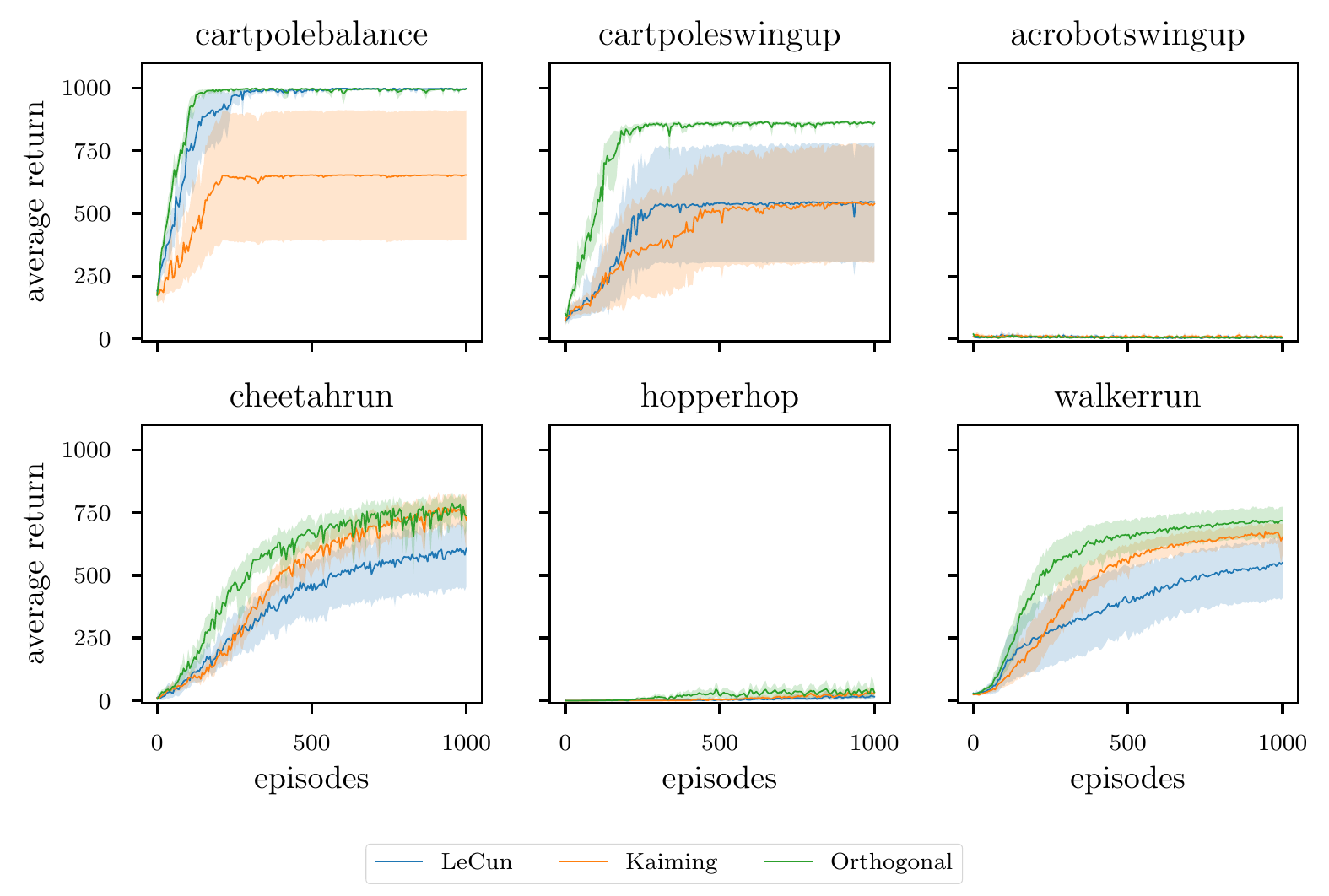}
    \caption{Mean performance of TD3 under LeCun, Kaiming and Orthogonal initialization.}
    \label{fig:TD3Init}
     \end{minipage}%
    \hfill%
   \begin{minipage}[t]{0.48\linewidth}
     \centering
    \includegraphics[width=\columnwidth]{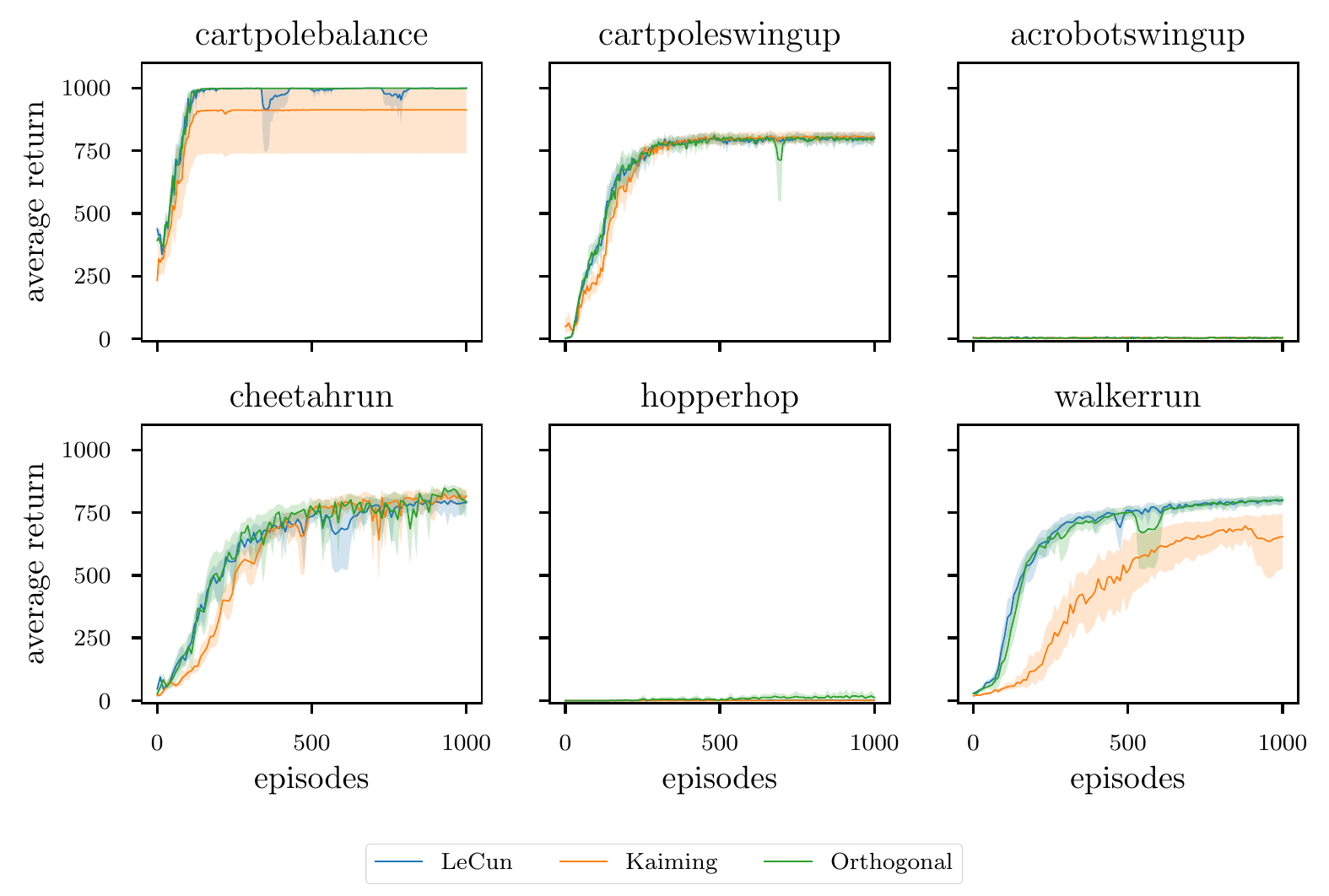}
    \caption{Mean performance of SAC under LeCun, Kaiming and Orthogonal initialization.}
    \label{fig:SACInit}
     \end{minipage}
\end{figure}

\begin{figure}
    \centering
    \includegraphics[width=0.85\columnwidth]{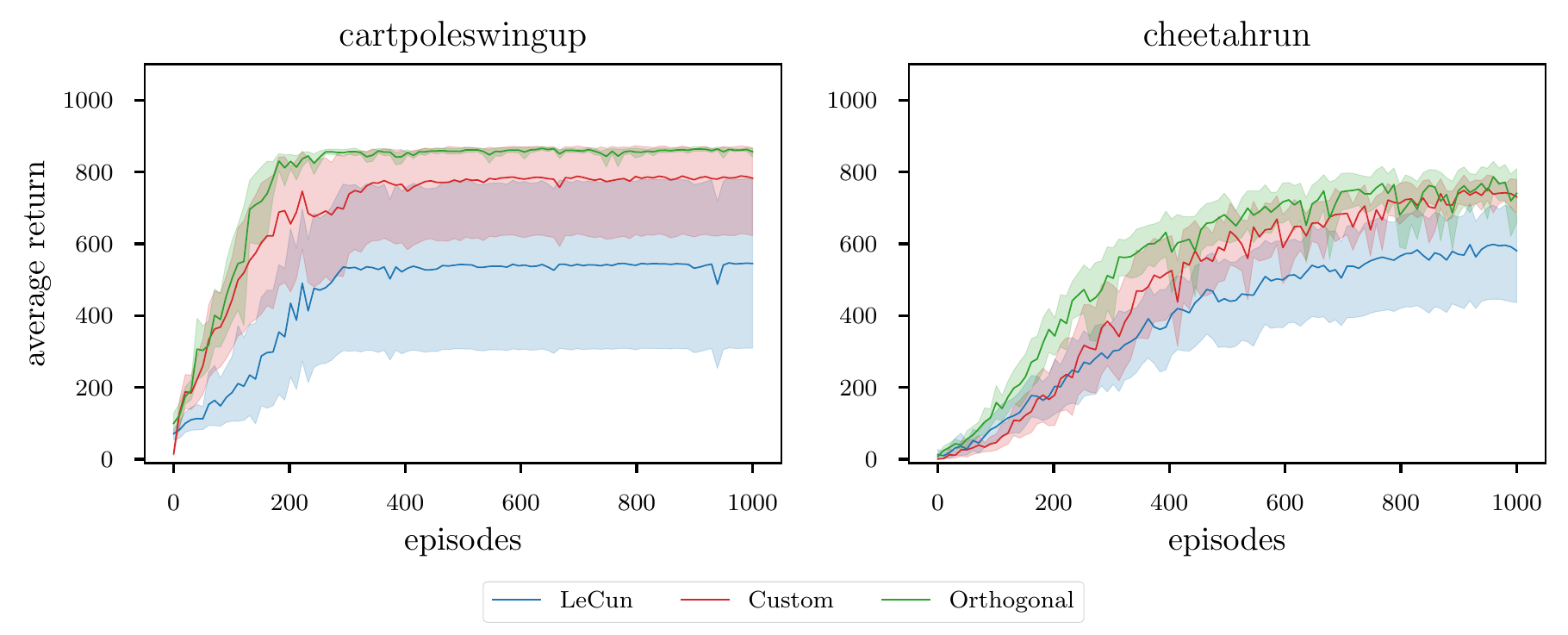}
    \caption{Replay buffer of TD3 initialized with 10,000 data points sampled from a uniform action distribution, to reduce the dependence on the initial policy parameters.}
    \label{fig:TD3LateStart}
\end{figure}

Figure~\ref{fig:histInit} shows the initial action distributions for the three policies under the different initialization schemes.
5000 states were randomly sampled from a standard normal distribution. Each density estimate was averaged over 100 random initializations. It shows that the mean spread increases with the different initializations, starting with the LeCun and reaching its maximum with Xavier/Kaiming initialization. With TD3's low noise level for exploration, the different initialization schemes greatly impact the initial action distribution. If the spread of the action distribution is larger than the action limits of the environment, the edges account for a lot of the probability mass due to action clipping. Heuristically, it is desirable to have a large spread while avoiding clipping by the environment to increase the diversity of actions explored. The experimental results on TD3 support these heuristics, as shown in figure~\ref{fig:TD3Init}.

Orthogonal initialization outperforms all other schemes in terms of learning speed and final performance. In more complex environments, Kaiming initialization only slows down the convergence, but the final performance is almost as good as with Orthogonal initialization. 
In the cartpole environment, the final performance is significantly worse.
This drop in performance is due to some of the runs using the Kaiming method, not resulting in any learning.  These runs weigh down the overall average performance. 
This finding indicates that the cartpole environment is more sensitive to reduced exploration.   

The original publication of TD3 \citep{fujimotoAddressingFunctionApproximation2018} suggests a custom procedure to reduce the impact of the initial parameters on the performance. This technique fills the replay buffer with a predefined number of transitions. The actions for those transitions are not sampled from the policy but a uniform distribution over the action space. This increases the performance and learning speed in our experiments but does not achieve similar performance levels as TD3 with orthogonal initialization. 
Figure~\ref{fig:TD3LateStart} shows how the proposed replay buffer initialization, in addition to the LeCun method, compares to only LeCun and Orthogonal schemes. The proposed technique claims to remove the dependency on the initial policy parameters. If this was the case, it should perform at least as good as the best initialization scheme.

The initial action distribution of SAC's stochastic policy is close to uniform. Only the Kaiming initialized version has more probability mass at the edges. This concentration of probability mass results in a downgrade of learning speed and the final performance shown in figure~\ref{fig:SACInit}.
The similar action distributions under LeCun and Orthogonal initialization result in almost identical learning behavior for all environments. This finding validates the hypothesis that the influence of initialization on the initial action distribution has a significant impact on learning behavior.

TRPO under different schemes shows similar results; we report these results in Appendix \ref{app:trpo}. Due to the high standard deviation at the start of training and the unbounded policy, most of the probability density is focused on the edges. The smaller the estimated means' variation, the smaller the overall variance of the initial action distribution will be. This, in turn, leads to more probability density for the actions in the center of the distribution. The initialization scheme that provides the least amount of spread to the means is LeCun. The performance of this initialization scheme is slightly better than that of Xavier and Orthogonal initialization. However, we do not observe a statistically significant effect, and experiments with more samples need to be conducted for confirmation. The difference in performance for the Xavier scheme is statistically significant, at least in cartpole-swingup and cheetah-run. 

\begin{figure}
   \begin{minipage}[t]{0.48\linewidth}
     \centering
    \includegraphics[width=\columnwidth]{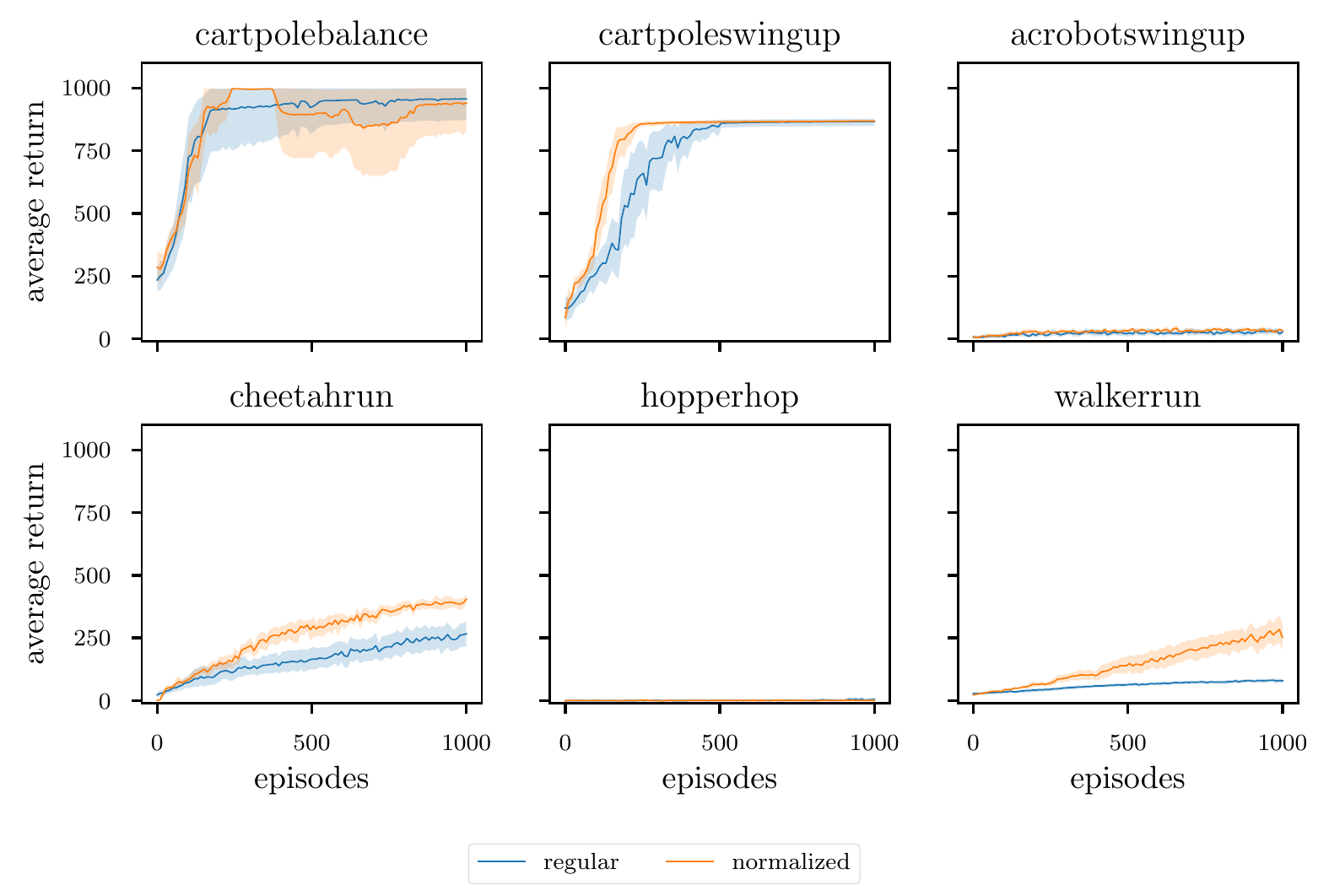}
    \caption{TRPO return with normalized vs. regular input states.}
    \label{fig:TRPONorm}
     \end{minipage}%
    \hfill%
   \begin{minipage}[t]{0.48\linewidth}
    \centering
    \includegraphics[width=\columnwidth]{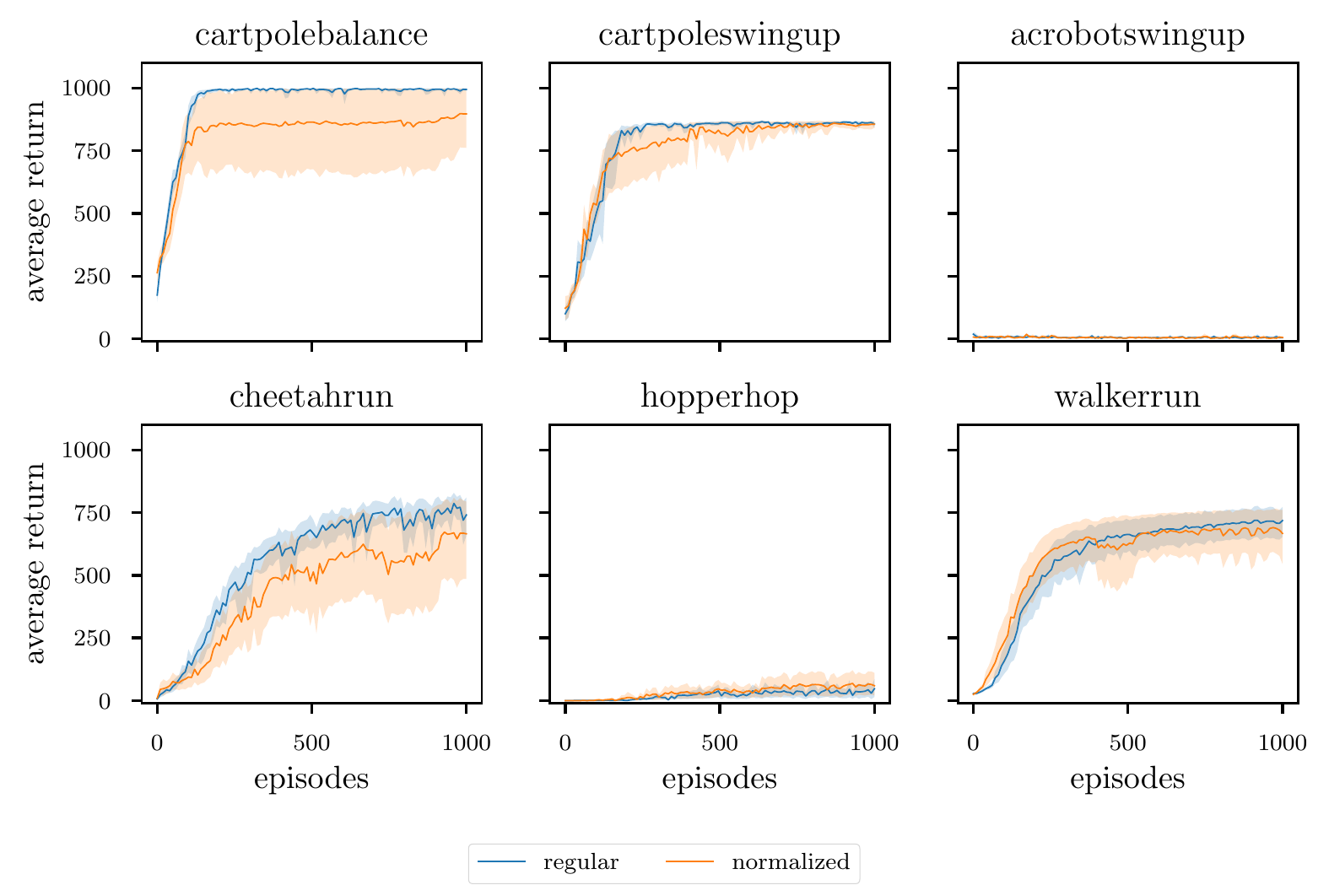}
    \caption{TD3 return with normalized vs. regular input states.}
    \label{fig:TD3Norm}
     \end{minipage}
\end{figure}

\subsection{Input Normalization}
Due to the conservative updates of TRPO, the algorithm converges very slowly. The slower convergence is especially evident in the performance of the more complex tasks, where the algorithm fails to converge in the given $1000$ episodes. With input normalization, the algorithm achieves significantly better performance in those tasks, as shown in figure~\ref{fig:TRPONorm}. 
Also, in the case of normalized inputs, the final performance of the high dimensional tasks is not even satisfactory. Furthermore, the confidence bounds indicate that the variance of the learning curve is reduced when normalization is employed.
This effect is not observable for TD3, where input normalization decreases the overall performance, as shown in figure~\ref{fig:TD3Norm}. Similar results were observed with SAC, which might indicate an incompatibility of input normalization with a replay buffer.  

\begin{figure}
   \begin{minipage}[t]{0.48\linewidth}
    \centering
    \includegraphics[width=\columnwidth]{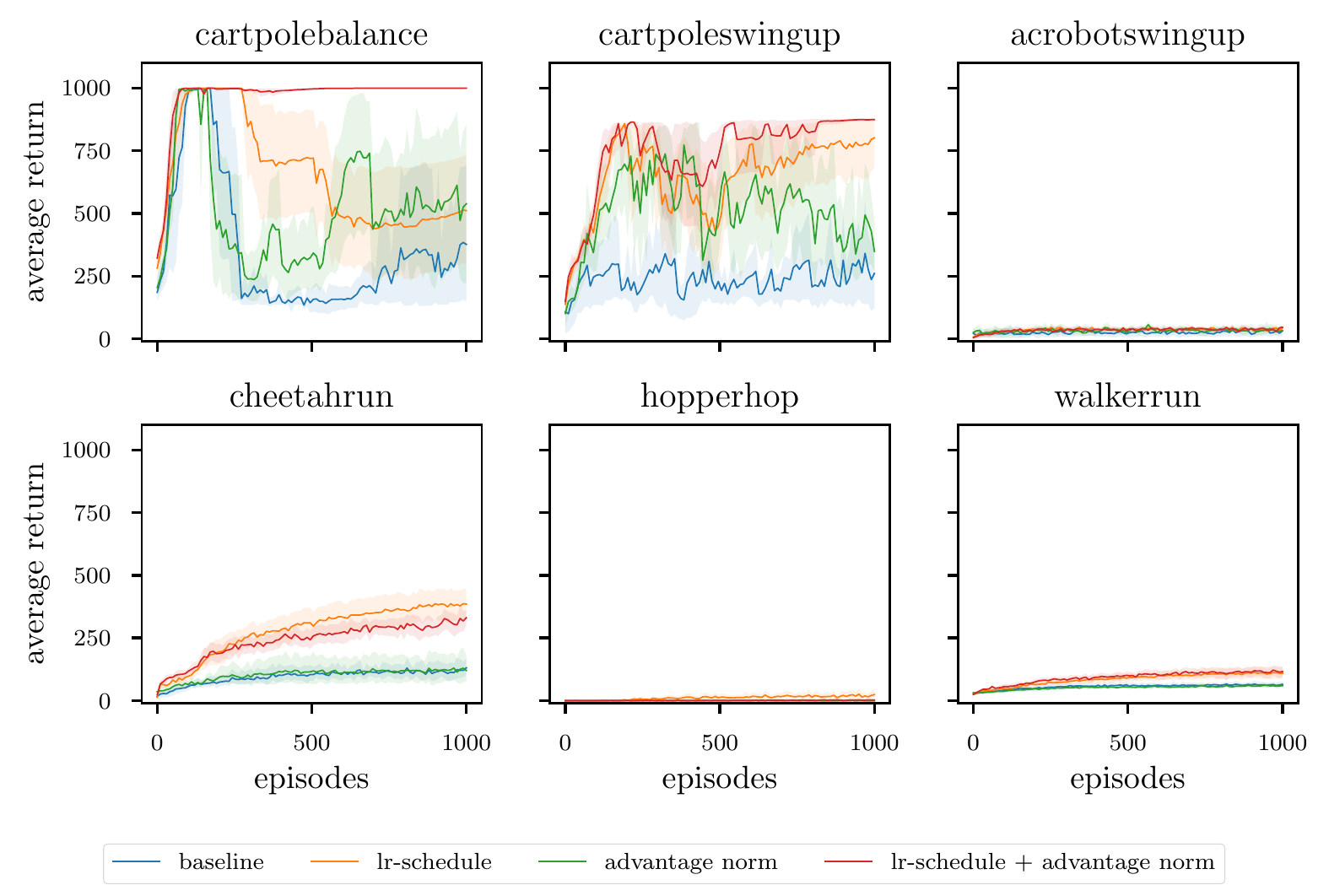}
    \caption{Baseline PPO performance compared with added Learning Rate Schedule and Advantage Normalization.}
    \label{fig:adaptive0}
     \end{minipage}%
    \hfill%
   \begin{minipage}[t]{0.48\linewidth}
    \centering
    \includegraphics[width=\columnwidth]{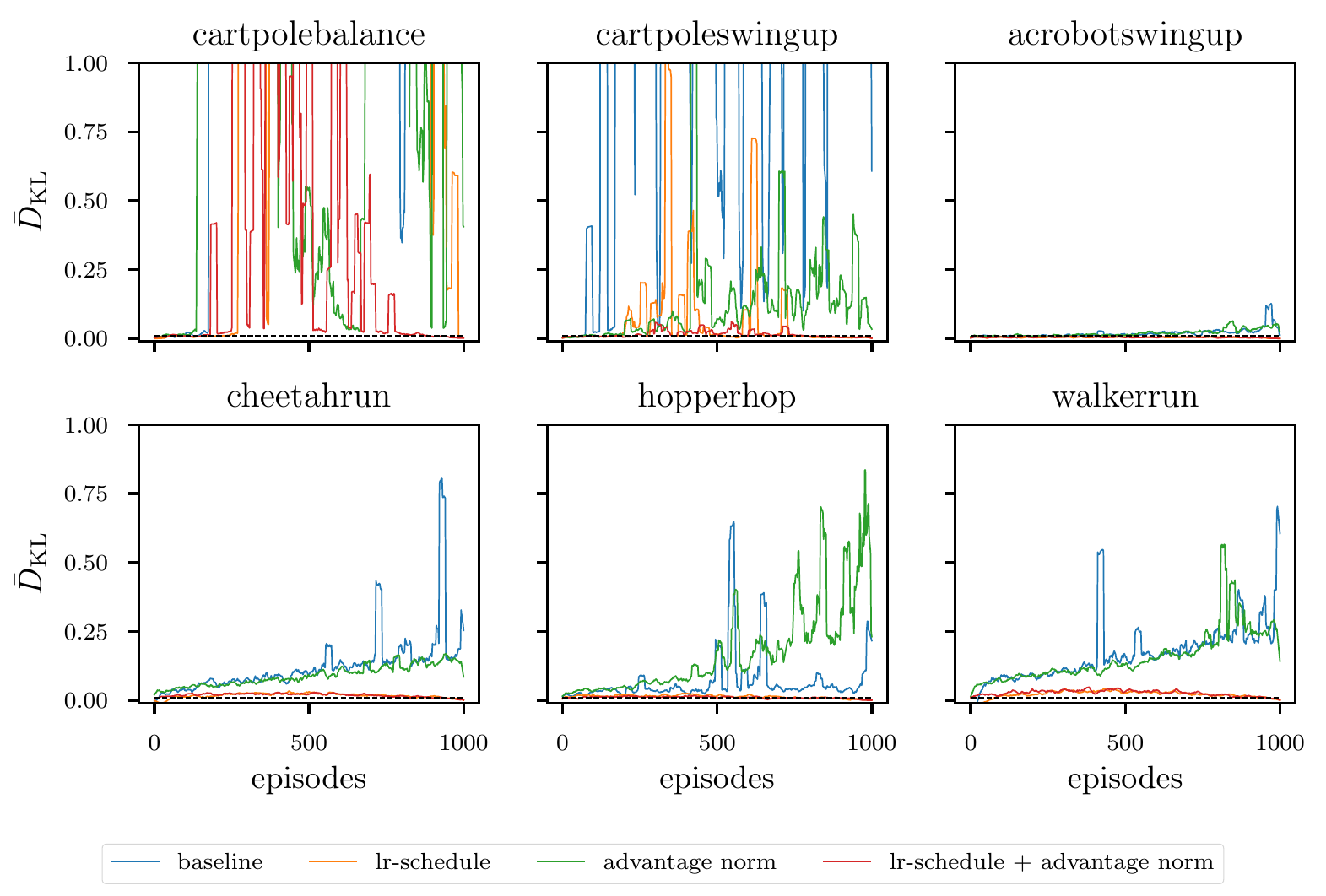}
    \caption{Mean KL-Divergence for PPO with Learning Rate Schedules and Advantage Normalization. The dashed line represents the desired trust region.}
    \label{fig:adaptive1KL}
     \end{minipage}
\end{figure}

\subsection{Adaptive Learning}
For experimental analysis of adaptive learning techniques, we chose to focus only on PPO because it relies on many of them and exhibits the most unstable learning behavior without using any of those techniques.
Figure~\ref{fig:adaptive0} shows how PPO performs in all environments if not aided by any learning adaptation. 

\begin{figure}
   \begin{minipage}[t]{0.48\linewidth}
    \centering
    \includegraphics[width=\columnwidth]{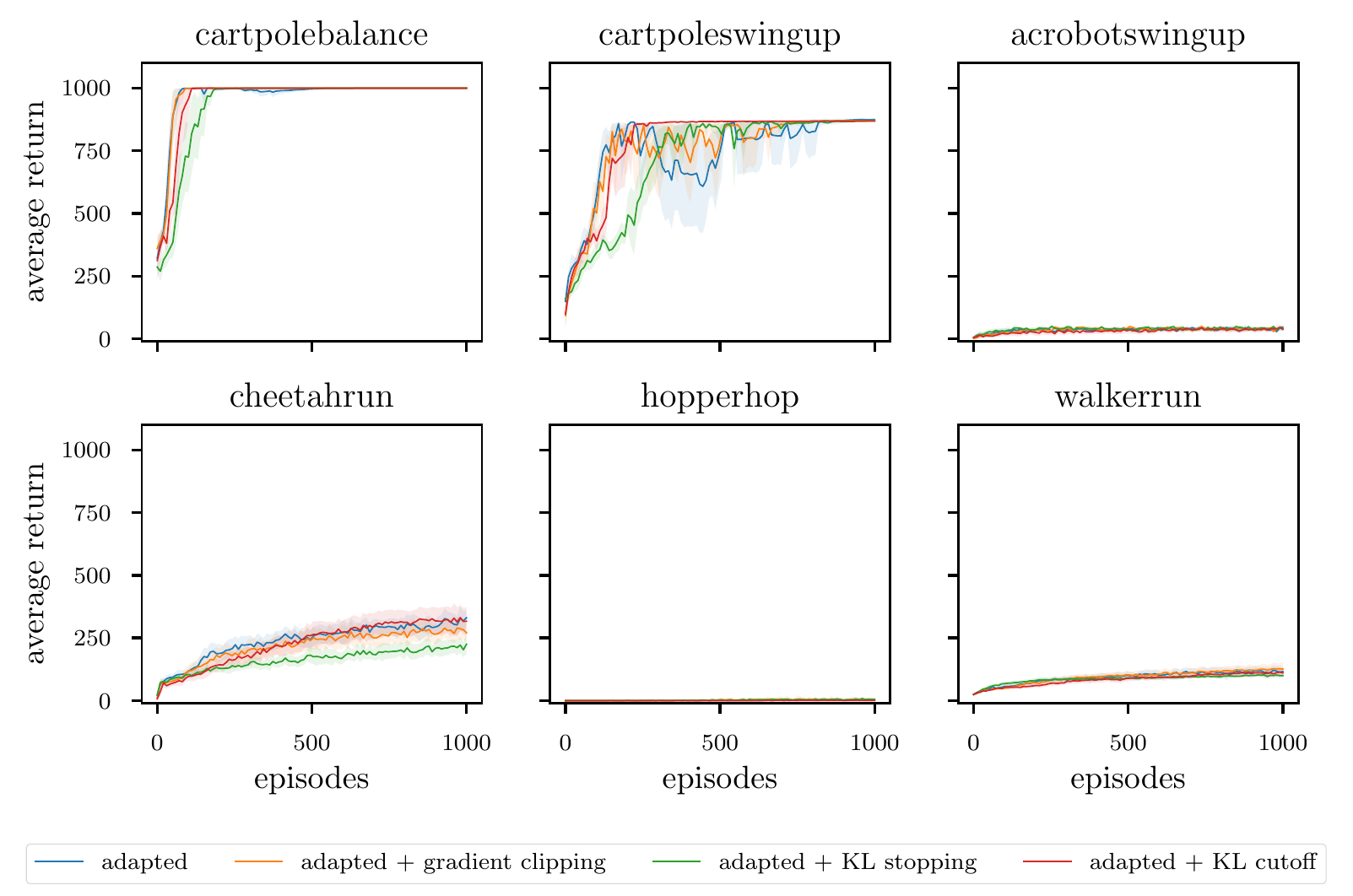}
    \caption{PPO performance with Learning Rate Schedules, Advantage Normalization and one additional technique.}
    \label{fig:adaptive2}
     \end{minipage}%
    \hfill%
   \begin{minipage}[t]{0.48\linewidth}
    \centering
    \includegraphics[width=\columnwidth]{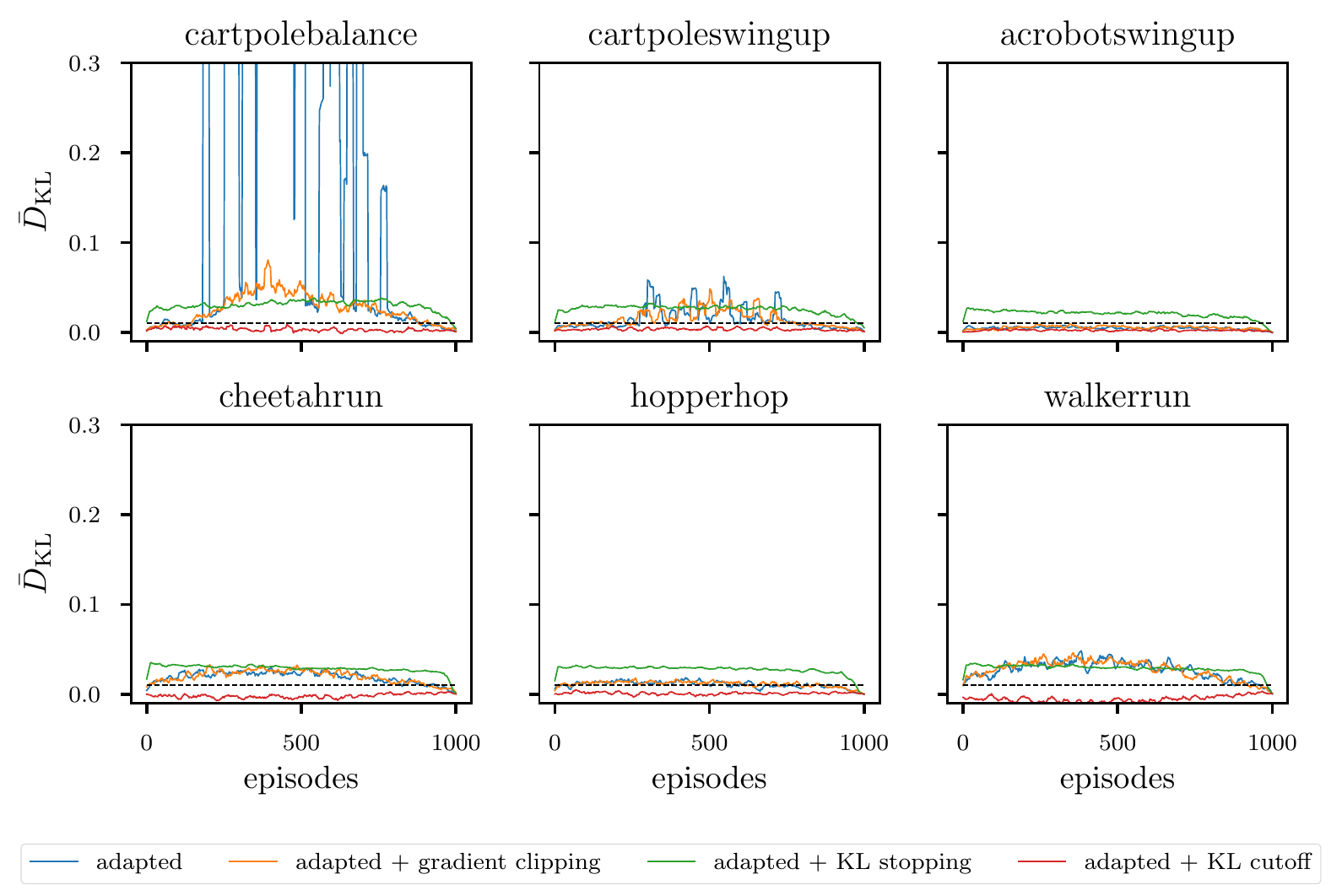}
    \caption{Mean KL-Divergence for PPO with LR Schedule, Advantage Normalization and an additional technique. The dashed line represents the desired trust region.}
    \label{fig:adaptive2KL}
     \end{minipage}
\end{figure}

The combination of LRS and AN are necessary to exhibit any learning progress for some environments. The KL-Divergence may explain why the vanilla implementation performs so poorly. Figure~\ref{fig:adaptive1KL} shows how the KL-Divergence of the vanilla implementation violates the desired trust region. Especially in the cartpole environment, large spikes are observable. LRS seems to reduce the average KL-Divergence but still allows for occasional spikes. Contrarily, AN does not reduce the average but filters the spikes. Using both techniques together achieves low KL-Divergence in all environments. The desired average KL-Divergence is commonly set to $0.01$, which seems reasonable for most environments except for cartpole-balance. One potential reason for this is that the closeness to linear dynamics might allow for more drastic policy changes.
Even with both adaptation techniques, the KL-Divergence is much higher than the desired value, and performance is very good.  
One of the remaining techniques can be added on top of LRS and AN to make sure this threshold is kept even tighter  (new baseline adapted). Figure~\ref{fig:adaptive2} shows the performance of the modified algorithms.

The figure shows that KL-Cutoff generally achieves performance as good as or better than the adapted baseline. However, it does reduce the learning speed slightly. 
KL-Stopping significantly reduces the learning speed and does not always provide better performance. Gradient Clipping does not seem to offer a significant benefit towards the adapted baseline. 
The resulting KL-Divergence is shown in figure~\ref{fig:adaptive2KL}. The only technique, which manages to uphold the desired threshold is KL-Cutoff. KL-Stopping even increases the average KL-Divergence by not allowing the clipped objective to optimize over enough epochs.
This is surprising since baselines like OpenAI Baselines\footnote[1]{\url{https://github.com/openai/baselines}} and OpenAI SpinningUp\footnote[2]{\url{https://github.com/openai/spinningup}} do not use this technique in their implementations. Only the original implementation\footnote[4]{\url{https://github.com/joschu/modular_rl}} deploys KL-Cutoff.
In general, it can be seen that without these techniques, PPO is unable to solve the benchmark tasks. State-of-the-art performance is achievable only with the right combination of methods. 

\section{Discussion and Conclusion}
We investigate the impact of initialization, normalization, and adaptive learning on state-of-the-art deep reinforcement learning algorithms through experimental methods. The results show that initialization changes the initial action distribution and therefore influences Deep RL algorithms differently than Deep SL algorithms. In deep RL, the initial action distribution dictates the exploration behavior, at least at the early stages. Thus, there is a need to develop RL specific initialization methods that account for the initial action distribution. Currently, Orthogonal initialization provides the best results and can be used for any architecture. 
Input normalization improved the performance of TRPO, but could not do so for TD3 and SAC. In general, our findings could indicate that input normalization should not be used with Q-learning algorithms. 
Finally, we investigate adaptive learning techniques applied to PPO. Our experiments show that the algorithm could not achieve state-of-the-art performance without such methods. We observe the best performance when using a combination of Learning Rate Scheduling, Advantage Normalization, and KL-Cutoff.

In general, Initialization and Adaptive Learning are implementation details that need to be carefully considered when implementing deep RL algorithms. Both can influence the performance and even cause algorithms not to learn at all. Overall, we conclude that implementation details have a strong influence on the final performance of an algorithm. Our findings encourage full transparency in RL research for the sake of reproducibility and reusability. It should become standard practice to include implementation details in publications if they affect performance significantly.

\section*{ACKNOWLEDGEMENTS}
We greatly acknowledge the funding of this work by Microsoft Germany and the Alfried Krupp von Bohlen und Halbach Foundation.



%



\bibliography{iclr2021_conference}
\bibliographystyle{iclr2021_conference}


\begin{appendices}
\newpage

\section{Additional Results for TRPO}\label{app:trpo}


\begin{figure}[!htb]
    \centering
    \includegraphics[width=\columnwidth]{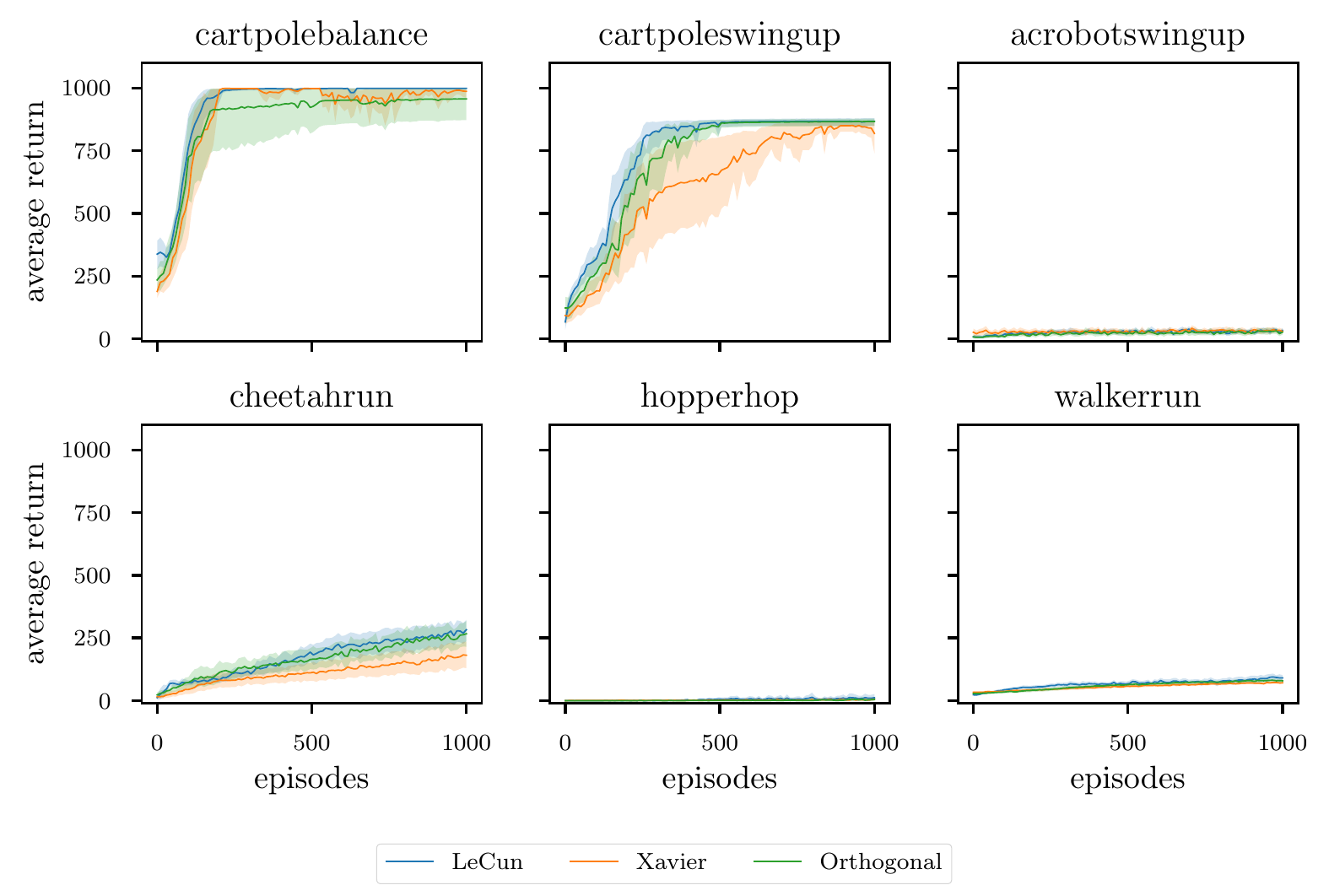}
    \caption{Mean performance of TRPO under LeCun, Xavier and Orthogonal initialization.}
    \label{fig:TRPOInit}
\end{figure}

\newpage
\section{Hyperparameters}
This section shows the default hyperparameters for the evaluated algorithms.
\begin{table}[H]
    \centering
    \scriptsize
    \begin{tabular}{p{0.3\linewidth} p{0.4\linewidth}}
      \toprule
        Parameter & Value \\
      \midrule
        Policy Network & (400, ReLU, 300, ReLU, Linear) \\
        Q-Network & (400, ReLU, 300, ReLU, Linear) \\
        Initialization & Orthogonal \\
        Learning Rate & 1e-3 \\
        Soft Target Update $\tau$ & 5e-3 \\
        Memory Size & 1e6 \\
        Batch Size & 100 \\
        Update Steps & 5 \\
        Discount Factor $\gamma$ & 0.99 \\
        Exploration Noise & 0.1 \\
        Target Policy Noise & 0.2 \\
        Policy Noise Clip & 0.5 \\
        Policy Update Frequency & 2 \\
      \bottomrule
    \end{tabular}
    \caption[TD3 Hyperparamters]
    {TD3 Hyperparamters}
\end{table}

\begin{table}[H]
    \centering
    \scriptsize
    \begin{tabular}{p{0.3\linewidth} p{0.4\linewidth}}
      \toprule
        Parameter & Value \\
      \midrule
        Policy Network & (64, Tanh, 64, Tanh, Linear) + Standard Deviation Variable \\
        Value-Network & (64, Tanh, 64, Tanh, Linear) \\
        Initialization & $w \sim \mathcal{U}\left[-\sqrt{\frac{1}{n}}, \sqrt{\frac{1}{n}}\right]$ \\
        Learning Rate & 1e-3 \\
        Batch Size & 5000 \\
        Discount Factor $\gamma$ & 0.99 \\
        GAE $\lambda$ & 0.95 \\
        Value Epochs & 80 \\
        Policy Epochs & 1 \\
        Trust Region $\delta$ & 0.01 \\
        CG Iterations & 10 \\
        CG Damping & 0.1 \\
        Backtrack Rate $\alpha$ & 0.8 \\
        Number of Backtracks $k$ & 10 \\
      \bottomrule
    \end{tabular}
    \caption[TRPO Hyperparamters]
    {TRPO Hyperparamters}
\end{table}

\begin{table}[H]
    \centering
    \scriptsize
    \begin{tabular}{p{0.3\linewidth} p{0.4\linewidth}}
      \toprule
        Parameter & Value \\
      \midrule
        Policy Network & (64, Tanh, 64, Tanh, Linear) + Standard Deviation Variable \\
        Value-Network & (64, Tanh, 64, Tanh, Linear) \\
        Initialization & $w \sim \mathcal{U}\left[-\sqrt{\frac{1}{n}}, \sqrt{\frac{1}{n}}\right]$ \\
        Learning Rate & 3e-4 \\
        Discount Factor $\gamma$ & 0.99 \\
        GAE $\lambda$ & 0.95 \\
        Batch Size & 2048 \\
        Mini Batch Size & 64 \\
        Epochs & 10 \\
        Clip Range $\epsilon$ & 0.2 \\
        Max KL-Divergence & 0.01 \\
        Entropy Coefficient & 0.1 \\
        KL-Cutoff Coefficient & 100 \\
        Max Gradient Norm & 0.5 \\
        Weight Decay & 0 / 1e-4 \\
      \bottomrule
    \end{tabular}
    \caption[PPO Hyperparamters]
    {PPO Hyperparamters}
\end{table}

\end{appendices}
\end{document}